\documentclass[letterpaper]{article}
\usepackage{helvet}
\usepackage{courier}
\usepackage[latin9]{inputenc}
\usepackage{array}
\usepackage{multirow}
\usepackage{graphicx}

\makeatletter

\pdfpageheight\paperheight
\pdfpagewidth\paperwidth

\providecommand{\tabularnewline}{\\}

\def\year{2017}\relax
\usepackage{aaai17}
\usepackage{times}
\nocopyright
  
\usepackage[pdftex,draft]{hyperref}

\makeatother

\begin{document}

\title{Depth CNNs for RGB-D scene recognition: learning from scratch better
than transferring from RGB-CNNs}

\author{Xinhang Song$^{1,2}$, Luis Herranz$^{1}$, Shuqiang Jiang$^{1,2}$\\
 $^{1}$Key Laboratory of Intelligent Information Processing of Chinese
Academy of Sciences (CAS), \\
Institute of Computing Technology, CAS, Beijing, 100190, China\\
 $^{2}$University of Chinese Academy of Sciences, Beijing, China\\
 \{xinhang.song, luis.herranz, shuqiang.jiang\}@vipl.ict.ac.cn}
\maketitle
\begin{abstract}
Scene recognition with RGB images has been extensively studied and
has reached very remarkable recognition levels, thanks to convolutional
neural networks (CNN) and large scene datasets. In contrast, current
RGB-D scene data is much more limited, so often leverages RGB large
datasets, by transferring pretrained RGB CNN models and fine-tuning
with the target RGB-D dataset. However, we show that this approach
has the limitation of hardly reaching bottom layers, which is key
to learn modality-specific features. In contrast, we focus on the
bottom layers, and propose an alternative strategy to learn depth
features combining local weakly supervised training from patches followed
by global fine tuning with images. This strategy is capable of learning
very discriminative depth-specific features with limited depth images,
without resorting to Places-CNN. In addition we propose a modified
CNN architecture to further match the complexity of the model and
the amount of data available. For RGB-D scene recognition, depth and
RGB features are combined by projecting them in a common space and
further leaning a multilayer classifier, which is jointly optimized
in an end-to-end network. Our framework achieves state-of-the-art
accuracy on NYU2 and SUN RGB-D in both depth only and combined RGB-D
data.
\end{abstract}

\section{Introduction}

Success in visual recognition mainly depends on the feature representing
the input data. Scene recognition in particular has benefited from
recent developments in the field. Most notably, massive image datasets
(ImageNet and Places\cite{Zhou2014}) provide the necessary amount
of data to train complex convolutional neural networks (CNNs)\cite{Krizhevsky2012,Simonyan2015},
with millions of parameters, without falling into overfitting. The
features extracted from models pretrained in those datasets have shown
to be generic and powerful enough to obtain state-of-the-art performance
in smaller datasets (e.g. MIT indoor 67\cite{Quattoni2009}, SUN397\cite{Xiao2010}),
just using an SVM\cite{Jeff2014} or fine-tuning, outperforming earlier
handcrafted paradigms (e.g. SIFT, HOG, bag-of-words).

Low cost depth sensors can capture depth information in addition to
RGB data. Depth can provide valuable information to model object boundaries
and understand the global layout of objects in the scene. Thus, RGB-D
models can improve recognition over mere RGB models. However, in contrast
to RGB data, which can be crowdsourced by crawling the web, RGB-D
data needs to be captured with a specialized and relatively complex
setup\cite{Silberman2011,Song2015a}. For this reason, RGB-D datasets
are orders of magnitude smaller than the largest RGB datasets, also
with much fewer categories. This prevents from training deep CNNs
properly, and handcrafted features are still a better choice for this
modality.

\begin{figure}
\begin{centering}
\includegraphics[width=0.14\columnwidth]{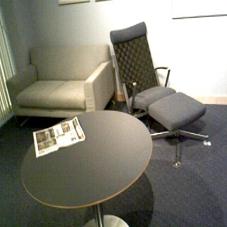}\includegraphics[width=0.14\columnwidth]{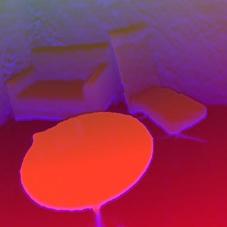}\textvisiblespace{}\includegraphics[width=0.14\columnwidth]{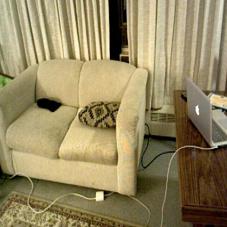}\includegraphics[width=0.14\columnwidth]{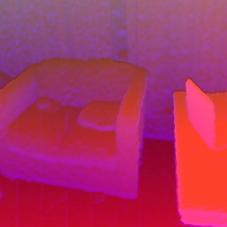}\textvisiblespace{}\includegraphics[width=0.14\columnwidth]{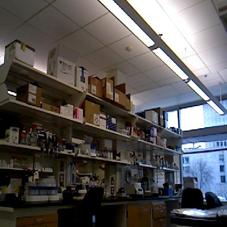}\includegraphics[width=0.14\columnwidth]{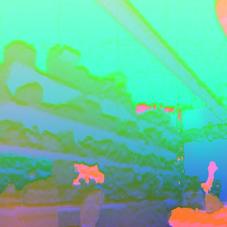}
\par\end{centering}
\begin{centering}
\includegraphics[width=0.9\columnwidth]{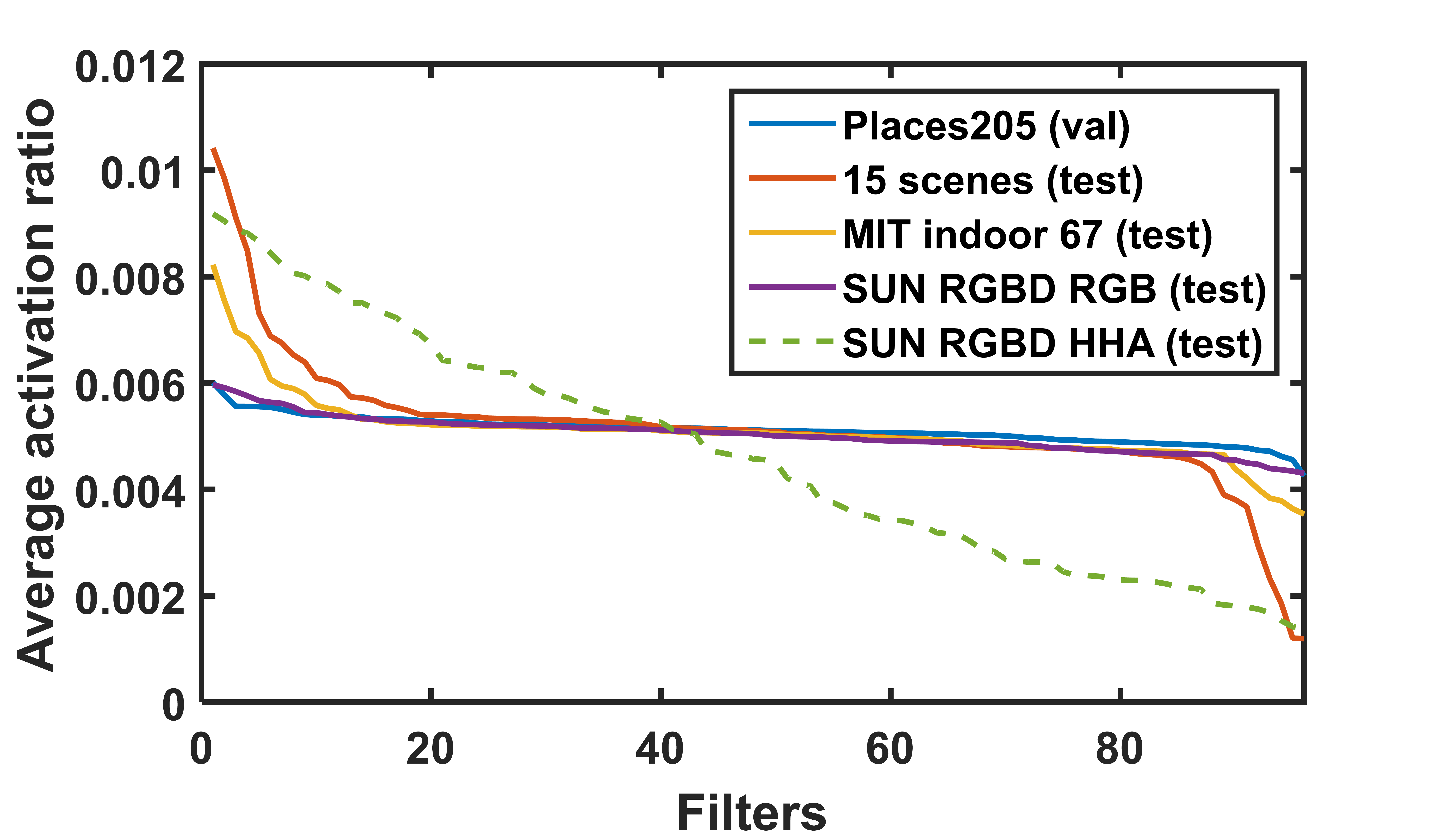}
\par\end{centering}
\begin{centering}
\includegraphics[width=0.95\columnwidth]{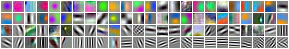}
\par\end{centering}
\caption{\label{fig:motivation}RGB and depth modalities. Top: examples of
scenes captured in RGB and depth (HHA encoding). Middle: average nonzero
activations of filter in the conv1 layer of Places-CNN for different
datasets. Bottom: Conv1 filters ordered by mean activation on SUN
RGB-D HHA.}

\vspace{-0.5cm}
\end{figure}

However, the recent SUN RGB-D dataset\cite{Song2015a} is significantly
larger than previous RGB-D scene datasets (e.g. NYU2\cite{Silberman2012}).
While still not large enough to train from scratch deep CNNs of comparable
size to RGB counterparts (10335 RGB-D images compared with 2.5 million
RGB images in Places), at least provides enough data for fine tuning
deep models (e.g. AlexNet-CNN on Places) without significant overfitting.
This approach typically exploit the HHA encoding for depth data\cite{GuptaECCV2014},
since it also a three channel representation (horizontal disparity,
height above ground, and angle with the direction of gravity, see
Figure~\ref{fig:motivation} top). Fine tuning is typically used
when the target dataset has limited data, but there is another large
dataset that covers a similar domain which can be exploited first
to train a deep model. Thus, transferring RGB features and fine tuning
with depth (HHA) data is the common practice to learn deep representations
for depth data\cite{Song2015a,Wang_2016_CVPR,Zhu_2016_CVPR,Gupta_2016_CVPR}.
However, although HHA images resemble RGB images and shapes and objects
can be identified, is it really reasonable reusing RGB features in
this inter-modal scenario?

In this paper we will focus on the low-level differences between RGB
and HHA data, and show that a large number of low-level filters are
either useless or ignored during fine tuning the network from RGB
to HHA. Figure~\ref{fig:motivation} (middle) shows the average activation
ratio (i.e. how often the activation is non-zero) of the 96 filters
in the layer conv1 of Places-CNN for different input data (sorted
in descending order). When a network is properly designed and trained,
it tends to show a balanced activation rate curve (e.g. conv1 activations
extracted from the Places validation set, where the curve is almost
a constant) meaning that most of the filters are contributing almost
equally to build discriminative representations. When transferred
to other RGB scene datasets, the curve is still very similar, showing
that the majority of the conv1 filters are still useful for 15 scenes
\cite{Lazebnik2006} and MIT Indoor 67. However, the curve for HHA
shows a completely different behavior, where only a subset of the
filters are relevant, while a large number are rarely activated, because
they are not useful for HHA (see Figure~\ref{fig:motivation} bottom).
Edges and smooth gradients are common, while, Gabor-like patterns
and high frequency patterns are seldom found in HHA data.

Thus, we observe that filters at the very bottom layers are crucial.
However, conventional full fine tuning from RGB CNNs can hardly reach
them (i.e. vanishing gradient problem), we explore other ways to make
better use of the limited data while focusing on bottom layers. In
particular, we compare the strategies of fine tuning only top and
only bottom layers, and propose a weakly supervised strategy to learn
filters directly from the data. In addition, we combine the pretrained
RGB and depth networks into a new network, fine tuned with RGB-D image
pairs. We show experimentally that these features lead to state-of-the-art
performance with depth and RGB-D, and provide some insights and evidences.
Code is available at \url{https://github.com/songxinhang/D-CNN}

\section{Related work}

\subsection{RGB-D scene recognition}

Earlier works use handcrafted features, engineered by an expert to
capture some specific properties considered representative. Gupta
\textit{et al. \cite{Gupta2015} }propose a method to detect contours
on depth images for segmentation, then further quantize the segmentation
outputs as local features for scene classification. Banica \textit{et
al. \cite{Banica_2015_CVPR} }quantize local features with second
order pooling, and use the quantized feature for segmentation and
scene classification. More recently, multi-layered networks can learn
features directly from large amounts of data. Socher \textit{et al.}\cite{Socher2012}
use a single layer CNN trained unsupervisedly on patches, and combined
with a recurrent convolutional network (RNN). Gupta \textit{et al.}\cite{GuptaECCV2014}
use R-CNN on depth images to detect objects in indoor scenes. Since
the training data is limited, they augment the training set by rendering
additional synthetic scenes.

Current state-of-the-art relies on transferring and fine tuning Places-CNN
to RGB and depth data\cite{Gupta_2016_CVPR,Wang_2016_CVPR,Zhu_2016_CVPR,Song2015a}.
Wang \textit{et al. }\cite{Wang_2016_CVPR} extract deep features
on both local regions and whole images, then combine the features
of all RGB and depth patches and images in a component aware fusion
method. Some approaches\cite{Zhu_2016_CVPR,Gupta_2016_CVPR} propose
incorporating CNN architectures to fine-tune jointly RGB and depth
image pairs. Zhu \textit{et al. }\cite{Zhu_2016_CVPR} jointly fine-tune
the RGB and depth CNN models by including a multi-model fusion layer,
simultaneously considering inter and intra-modality correlations,
meanwhile regularizing the learned features to be compact and discriminative.
Alternatively, Gupta \textit{et al. \cite{Gupta_2016_CVPR} }propose
to transfer RGB CNN model to the depth data according to the RGB and
depth image pairs. 

In this paper we avoid relying on large yet biased RGB models to obtain
depth features, and train depth CNNs using weak supervision directly
from depth data, learning truly depth-specific and discriminative
features, compared with those transferred and adapted from biased
RGB models.

\subsection{Weakly-supervised CNNs}

Recently, several works propose weakly supervised frameworks\cite{Durand_2016_CVPR,Bilen_2016_CVPR,Oquab_2015_CVPR},
specially for object detection (object labels are known but not the
bounding boxes). Oquab \textit{et al. }\cite{Oquab_2015_CVPR} propose
an object detection framework to fine tune pretrained CNNs with multiple
regions, where a global max-pooling layer selects the regions to be
used in fine tuning. Durand \textit{et al. }\cite{Durand_2016_CVPR}
extend this idea by selecting both useful (positive) and ``useless''
(negative) regions with a maximum and minimum mixed pooling layer.
The weakly supervised detection network in \cite{Bilen_2016_CVPR}
uses a region proposal method to select regions.

These works rely on CNNs already pretrained on large datasets, and
weak supervision is used in a subsequent fine tuning or adaptation
stage to improve the final features for a particular task. In contrast,
our motivation is training when data is very scarce, with a weakly
supervised CNN that does not rely on any pretrained CNNs. In fact
it is used to pretrain convolutional layers prior to fine tuning with
full images.

\begin{figure}
\begin{centering}
\begin{tabular}{ccc}
\multicolumn{3}{c}{\includegraphics[width=0.9\columnwidth]{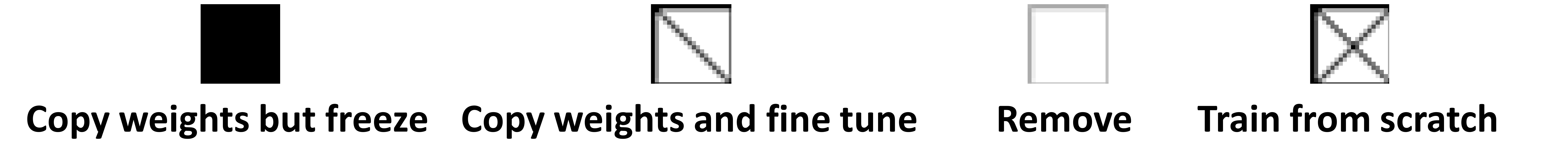}}\tabularnewline
\includegraphics[height=2.5cm]{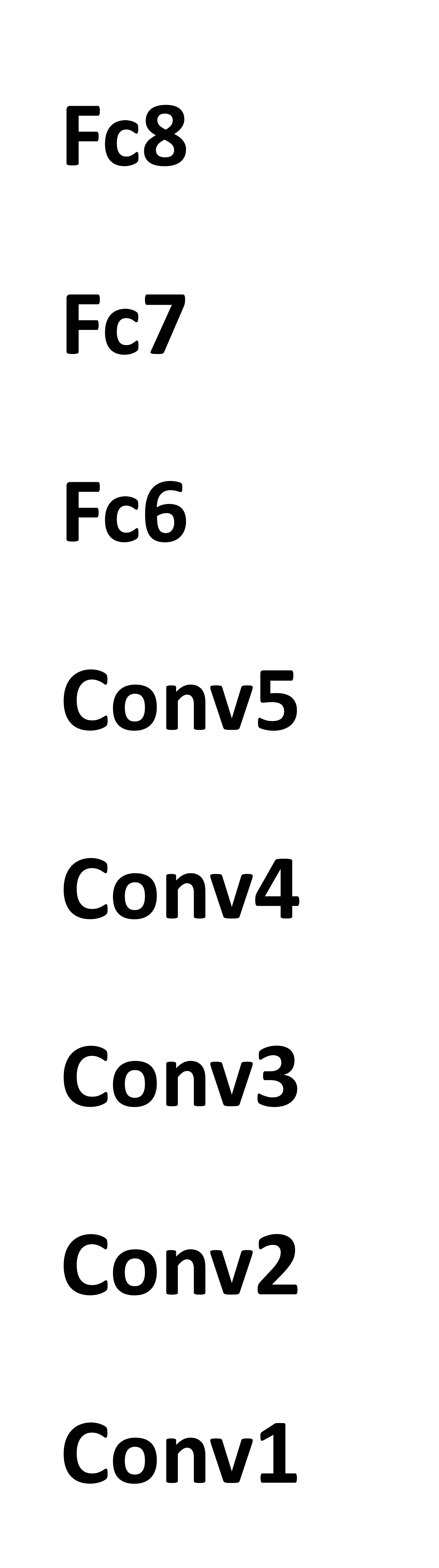}\includegraphics[height=2.5cm]{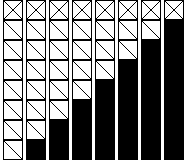} & \includegraphics[height=2.5cm]{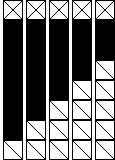} & \includegraphics[height=2.5cm]{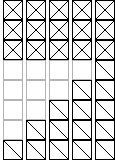}\tabularnewline
(a) FT-top & (b) FT-bottom & (c) FT-keep\tabularnewline
\end{tabular}
\par\end{centering}
\caption{\label{fig:ft-strategy}Strategies to fine tune Places-CNN withdepth
data, (a) only top layers, bottom layers are frozen, (b) only bottom
layers, top layers are frozen; (c) bottom layers (top layers retrained
and some convolutional layers removed). Each column represents a particular
setting.}
\end{figure}

\section{Transferring from RGB to depth}

\paragraph{Fine tuning Places-CNN with depth data}

Transferring RGB CNNs and fine tuning with RGB data (intra-modal transfer)
has been well studied. In general, low-level filters from bottom layers
capture generic patterns from the real visual world, and can be reused
effectively in datasets with the same modality and similar characteristics.
Thus, fine tuning one or two top layers (e.g. fc6, fc7) is often enough.
These layers are more dataset-specific and need to be rewired to the
new target categories\cite{Pulkit_2014}. In contrast, RGB and depth
images have significantly different low-level visual patterns and
regularities (e.g. depth patterns in HHA encoding are typically smooth
variations, contrasts and borders, but without textures and high frequency
patterns). Since the bottom convolutional layers are essential to
capture this modality-specific visual appearances and regularities,
only fine tuning top layers seem insufficient to adapt the RGB CNN
properly to depth data. 

Since we want to focus on bottom layers, we compare conventional fine
tuning with other strategies that reach bottom layers better (see
Fig.~\ref{fig:ft-strategy}, each column represents a particular
setting). The strategies differ basically on which layers are fine
tuned, trained and which remain unaltered. Using the AlexNet architecture
and the pretrained Places-CNN, we first compare three strategies:
a) \textit{FT-top (Places-CNN),} the conventional method where only
a few top layers are fine tuned, b)\textit{ FT-bottom (Places-CNN),}
where a few bottom layers are frozen, and c) \textit{FT-keep (Places-CNN)},
top layers are directly removed. Note that fc8 is always trained,
since it must be resized according to the target number of categories. 

\begin{figure}
\begin{centering}
\includegraphics[bb=170bp 0bp 2000bp 950bp,clip,width=0.95\columnwidth]{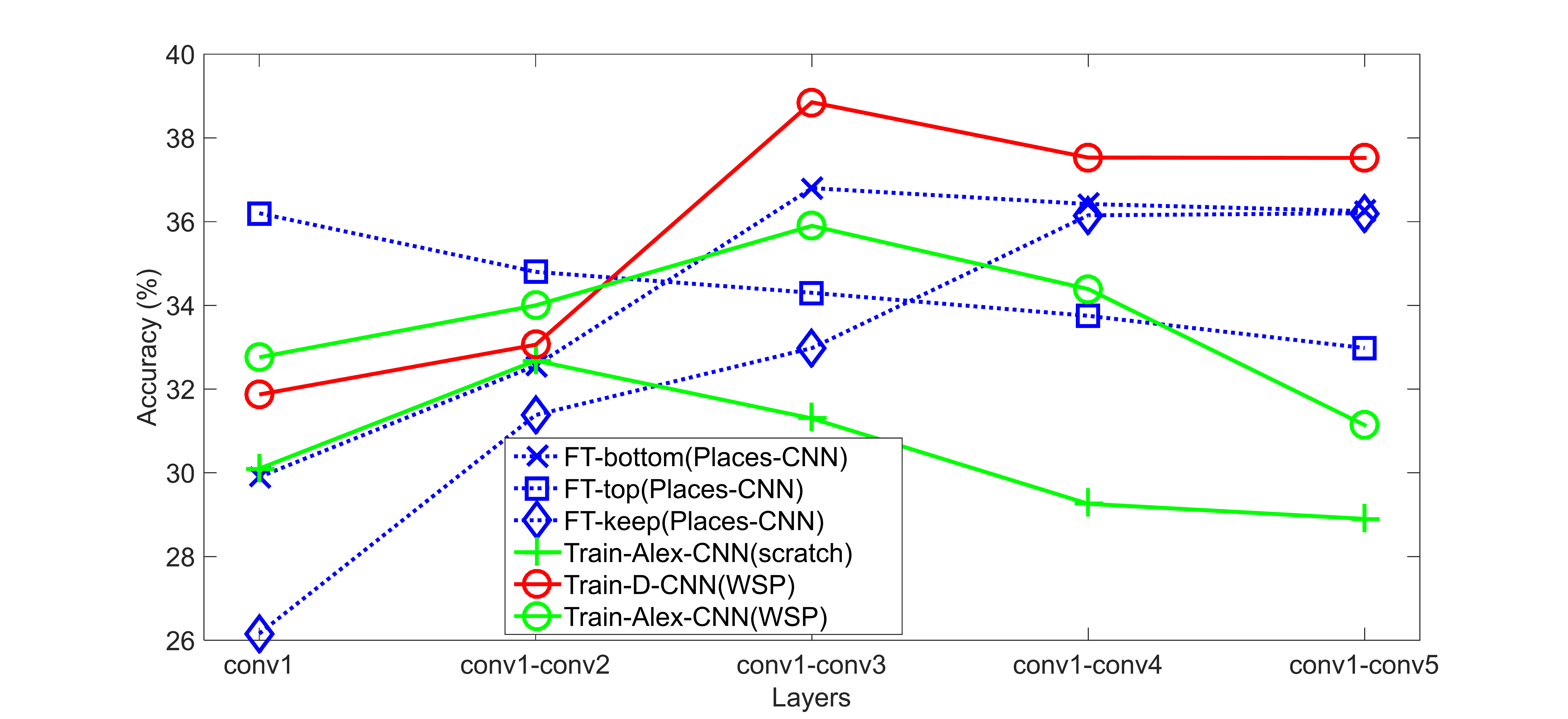}
\par\end{centering}
\caption{\label{fig:Comparisons-between-different}Comparisons of different
learning strategies on depth images in accuracy (\%) obtained with
softmax, the fine tuning strategies are illustrated in Fig.~\ref{fig:ft-strategy}
and training strategies are in Fig.~\ref{fig:tfs-strategy}.}

\vspace{-0.5cm}
\end{figure}

The classification accuracy on depth data (from SUN RGB-D dataset)
with different strategies is shown in Fig.~\ref{fig:Comparisons-between-different}.
Fine tuning top layers (\textit{FT-top}) does not help significantly
until including bottom convolutional layers, which is the opposite
to fine tuning for RGB\cite{Pulkit_2014}, where fine tuning one or
two top layers is almost enough to reach the maximum gain, and further
extending to bottom layers helps very marginally. In contrast, fine
tuning only the three bottom layers (\textit{FT-bottom}) achieves
36.5\% which is higher than fine tuning all layers. Furthermore, fine
tuning after removing top layers (\textit{FT-keep}) is also comparable
to fine tuning all layers. All these results support our intuition
that bottom layers are much more important than top layers when transferring
RGB to depth, and that conventional transfer learning and adaptation
tools used in RGB modality may not be effective in this inter-modal
case.

\paragraph{More insight from \textit{conv1} layer}

\begin{figure}
\begin{centering}
\setlength{\tabcolsep}{2pt} \renewcommand{\arraystretch}{1}%
\begin{tabular}{cccc}
\includegraphics[width=0.23\columnwidth]{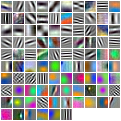} & \includegraphics[width=0.23\columnwidth]{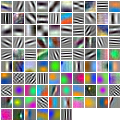} & \includegraphics[width=0.23\columnwidth]{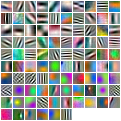} & \includegraphics[width=0.23\columnwidth]{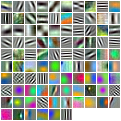}\tabularnewline
(a) & (b) & (c) & (d)\tabularnewline
\includegraphics[width=0.23\columnwidth]{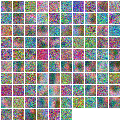} & \includegraphics[width=0.23\columnwidth]{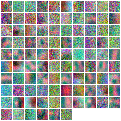} & \includegraphics[width=0.23\columnwidth]{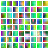} & \includegraphics[width=0.23\columnwidth]{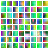}\tabularnewline
(e) & (f) & (g)  & (h)\tabularnewline
\end{tabular}
\par\end{centering}
\caption{\label{fig:Visual_conv1}Visualizing the first convolutional layer
(conv1): (a) Places-CNN; (b) full fine tuned Places-CNN; (c) FT-bottom
(Places-CNN); (d) FT-keep (Places-CNN), conv1; (e) Train-Alex-CNN
(scratch); (f) Train-Alex-CNN (WSP), training with patches ($99\times99$
pixels); (g) WSP-CNN, kernel size $5\times5$ pixels, training with
patches ($35\times35$ pixels); (f) Train-D-CNN (WSP). All methods
are trained/fine tuned using only the depth data from SUN RGB-D. }

\vspace{-0.5cm}
\end{figure}

To provide complementary insight of why bottom layers are more important
we focus on the filters from the first convolutional layer, i.e. conv1,
shown in Fig.~\ref{fig:Visual_conv1}. Although there is some gain
in accuracy, it can be observed that only a few particular filters
have noticeable changes during the fine tuning process. This suggest
that the CNN is reusing RGB filters, and thus trying to find RGB-like
patterns in depth data. Additionally, Fig.~\ref{fig:motivation}
middle shows that a large number of filters from Places-CNN are significantly
underused on depth data (while they are properly used on RGB data).
These observations suggest that reusing Places-CNN filters for conv1
and other bottom layers may not be a good idea. Moreover, since filters
also represent tunable parameters, this results in a model with too
many parameters that is difficult to train with limited data. 

\section{Weakly supervised pretrained CNN}

\begin{figure*}[!t]
\begin{centering}
\includegraphics[bb=60bp 100bp 810bp 510bp,clip,width=0.9\textwidth]{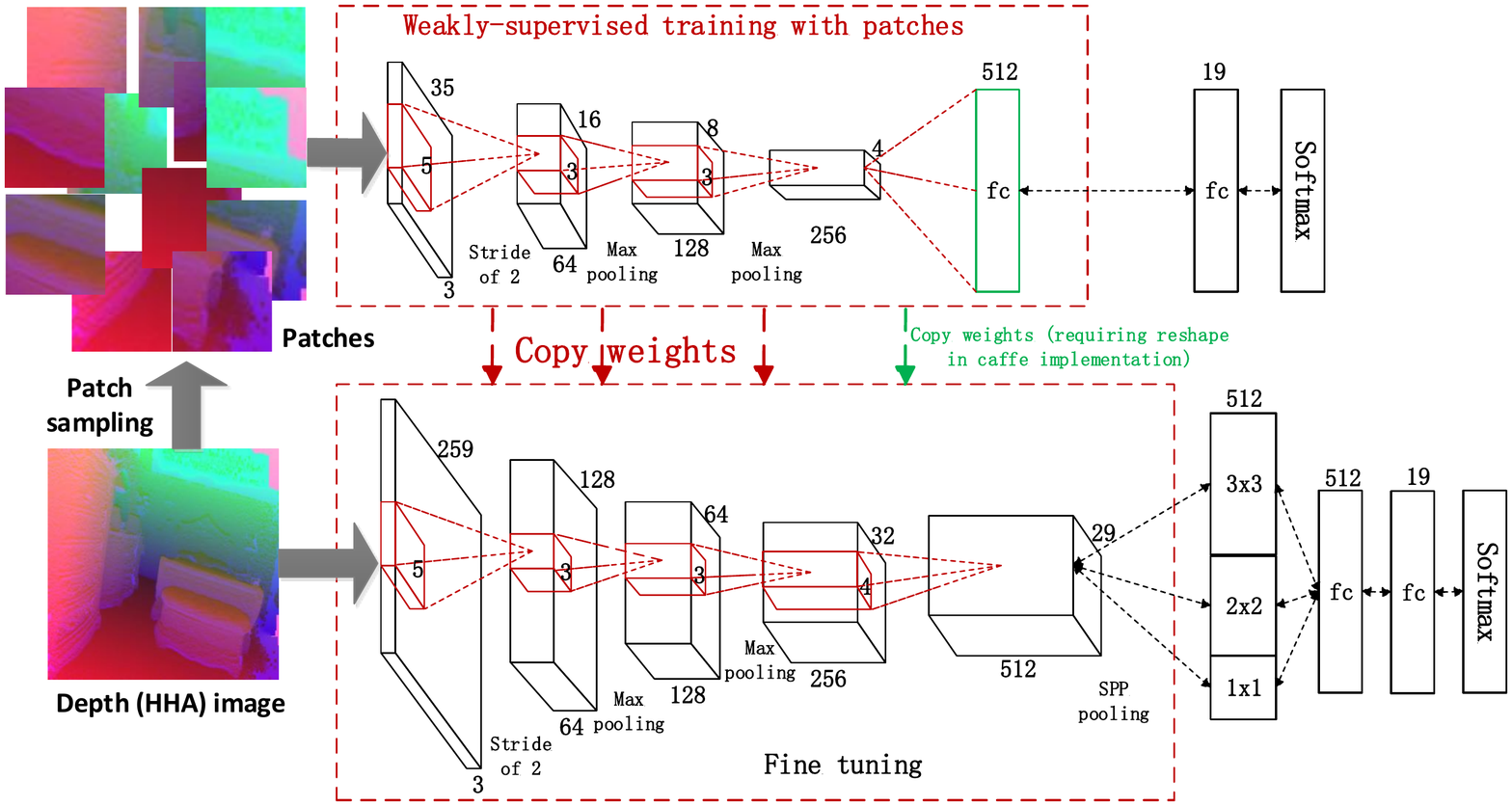}
\par\end{centering}
\caption{\label{fig:RGB-D-scene-recognition}Two-step learning of depth CNNs
combining weakly supervised pretraining and fine tuning.}

\vspace{0cm}
\end{figure*}
 
\begin{figure}[h]
\begin{centering}
\begin{tabular}{ccc}
\multicolumn{3}{c}{\includegraphics[width=0.9\columnwidth]{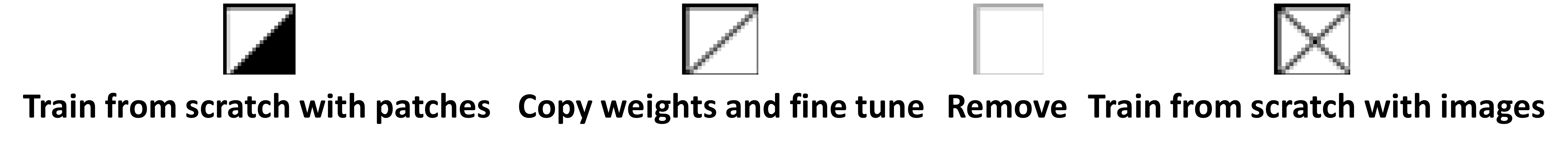}}\tabularnewline
\includegraphics[height=2.8cm]{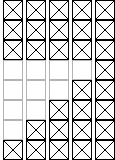} & \includegraphics[height=2.8cm]{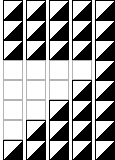} & \includegraphics[height=2.8cm]{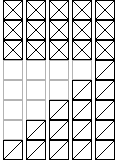}\tabularnewline
(a) Scratch & (b) WSP & (c) FT (WSP)\tabularnewline
\end{tabular}
\par\end{centering}
\caption{\label{fig:tfs-strategy}Training strategies for Alex-CNN variants
with depth images, (a) from scratch, (b) weakly-supervised with patches,
and (c) fine-tuned after weakly supervised trainig with patches.}
\vspace{0cm}
\end{figure}

It is difficult to learn deep CNN from scratch with depth images,
due to the lack of enough training data. We alternatively train CNN
from scratch with different architectures (see Fig.~\ref{fig:tfs-strategy}a),
from shallow to deep. The visualization of conv1 (architecture with
two convolutional layers) is illustrated in the top row of Fig.~\ref{fig:Visual_conv1}e.
However, we cannot observe any visual regularities from the visualizations,
even though this is a more shallow network. 

So in order to adapt the amount of training data and the complexity
(i.e. number of parameters) of the model, we modify slightly the training
procedure. First, we reduce the support of the CNN by working on patches
rather than on the whole image. Thus networks with simpler architectures
with fewer trainable parameters are capable. Additionally, we can
extract multiple patches from a single image, increasing the amount
of training data. Second, we include weakly supervised training. Patches
typically cover objects or parts, not the whole scene. However, since
we do not have those local labels we use the scene category as weak
label, since we know that all patches in a given image belong to the
same scene. We refer to this network as weakly supervised patch-CNN
(WSP-CNN).

\subsection{Weak supervision on patches}

The weakly-supervised strategy can be used for Alex-CNN training.
We first sample a grid of $4\times4$ patches with $99\times99$ pixels
for weakly-supervised pretraining, and fine tune it with full images.
When switching the architecture from WSP-CNN to full Alex-CNN, the
amount of connections in fc6 changes. Thus, only the weights of the
convolutional layers of the pretrained WSP-CNN are copied for fine
tuning, similarly as other weakly supervised methods\cite{Durand_2016_CVPR,Bilen_2016_CVPR}.
Fig.~\ref{fig:Comparisons-between-different} shows that using weak
supervision on patches (WSP) significantly outperforms training with
full image (compare \textit{Train-Alex-CNN (WSP)} vs \textit{Train-Alex-CNN
(scratch)}). Furthermore, in the conv1 filters shown in Fig.~\ref{fig:Visual_conv1}f
(WSP) the depth specific-patterns are much more evident than in Fig.~\ref{fig:Visual_conv1}e
(full image). Nevertheless, they still show a significant amount of
noise, which suggests that AlexNet is still too complex, and perhaps
the kernels may be too big for depth data.

Since the complexity of depth images is significantly lower than that
of RGB images (e.g. no textures), we reduced the size of the kernels
in each layer to train our WSP-CNN, which consists of C (C=3 works
best after evaluation) convolutional layers (see Fig~\ref{fig:RGB-D-scene-recognition}
top row for the architecture detail). The sizes of the kernelsare
$5\times5$ (stride 2), $3\times3$ and $3\times3$, and the size
of max pooling is $2\times2$, stride 2. We sample a grid of $7\times7$
patches with $35\times35$ pixels for weakly-supervised training.

\subsection{Depth-CNN}

Fig~\ref{fig:RGB-D-scene-recognition} bottom shows the full architecture
of our depth-CNN (D-CNN). After training the WSP-CNN, we transfer
the weights of the convolutional layers. The output of conv4 in D-CNN
is $29\times29\times512$, almost 50 times larger than the output
of pool5 (size of $6\times6\times256$) in Alex-CNN, which leads to
50 times more parameters in this part. In order to reduce the parameters,
we include three spatial pyramid pooling (SPP) \cite{He_2014} layers
with size of $29\times29$,$15\times15$, $10\times10$. SPP also
captures spatial information and allows us to train the model end-to-end.
The proposed \textit{Train-D-CNN (WSP) }outperforms both fine tuning
and weakly-supervised training of Alex-CNN (see in Fig.~\ref{fig:Comparisons-between-different}).
Comparing the visualizations in Fig.~\ref{fig:Visual_conv1}, the
proposed WSP-CNN and D-CNN learn more representative kernels, supporting
the better performance. This also suggests that a smaller kernel size
is suitable for depth data.

\section{RGB-D fusion}

Most previous works use two independent networks for RGB and depth,
optimizing in different independent stages the fusion parameters,
fine tuning and classification\cite{Zhu_2016_CVPR,Wang_2016_CVPR,Wang2015}.
In contrast, we integrate both RGB-CNN, depth-CNN and the fusion procedure
into an integrated RGB-D-CNN, which can be trained end-to-end, jointly
learning the fusion parameters and fine tuning both RGB layers and
depth layers of each branch. As fusion mechanism, we use a network
with two fully connected layers followed by the loss (i.e. fusion
network), on top of the concatenated feature $\left[F_{rgb},F_{d}\right]$,
where $F_{rgb}\in R^{D_{r}\times1}$ and $F_{depth}\in R^{D_{d}\times1}$
are the RGB and depth features, respectively. The first layer of the
fusion network learns modality-specific projections $\bar{W}=\left[W_{rgb},W_{depth}\right]$
to a common feature space as 
\begin{equation}
F_{rgbd}=\left[W_{rgb}\:W_{depth}\right]\left[\begin{array}{c}
F_{rgb}\\
F_{depth}
\end{array}\right]=\bar{W}\left[\begin{array}{c}
F_{rgb}\\
F_{depth}
\end{array}\right]
\end{equation}
where $W_{rgb}\in R^{D_{rgbd}\times D_{r}}$ and $W_{depth}\in R^{D_{rgbd}\times D_{d}}$
are the modality-specific projection matrices.

Recent works exploit metric learning\cite{Wang2015}, Fisher vector\cite{Wang_2016_CVPR}
and correlation analysis\cite{Zhu_2016_CVPR} to reduce the redundancy
in the joint RGB-D representation. It is important to note that this
step is particularly effective when RGB and depth features are significantly
correlated. This is likely to be the case in recent works when both
RGB and depth feature extractors are fine tuned versions of the same
CNN model (e.g. Places-CNN). In our case depth models are learned
directly from the data and independently from RGB, so they are already
much less correlated, even without explicit multi-modal analysis.

\begin{table}
\caption{\label{tab:Comparisons-of-different}Ablation study for different
models (accuracy \%).}
\centering{}%
\begin{tabular}{c|cc|cc|c}
\hline 
Arch. & \multicolumn{4}{c|}{Alex-CNN} & D-CNN\tabularnewline
\hline 
Weights & \multicolumn{2}{c|}{Places-CNN} & \multicolumn{2}{c|}{Scratch} & Scratch\tabularnewline
\hline 
Layer & - & FT & Train & WSP & WSP\tabularnewline
\cline{2-6} 
pool1 & 17.2 & 20.3 & 22.3 & 23.5 & \textbf{25.3}\tabularnewline
pool2 & 25.3 & 27.5 & 26.8 & 30.4 & \textbf{33.9}\tabularnewline
conv3 & 27.6 & 29.3 & 29.8 & \textbf{35.1} & 34.6\tabularnewline
conv4 & 29.5 & 32.1 & - & - & \textbf{38.3}\tabularnewline
pool5 & 30.5 & 35.9 & - & - & -\tabularnewline
fc6 & 30.8 & 36.5 & 30.7 & 36.1 & -\tabularnewline
fc7 & 30.9 & 37.2 & 32.0 & 36.8 & \textbf{40.5}\tabularnewline
fc8 & - & 37.8 & 32.8 & 37.5 & \textbf{41.2}\tabularnewline
\hline 
\end{tabular}\vspace{-0.5cm}
\end{table}

\section{Experiments}

\paragraph{Dataset}

We evaluate our approach in two datasets: NYU depth dataset second
version (NYUD2) \cite{Silberman2012} and SUN RGB-D\cite{Song2015a}.
The former is a relatively small dataset with 27 indoor categories,
but only a few of them are well represented. Following the split in
\cite{Silberman2012}, all 27 categories are reorganized into 10 categories,
including 9 most common categories and an 'other' category consisting
of the remaining categories. The training/test split is 795/654 images.
SUN RGB-D contains 40 categories with 10335 RGB-D images. Following
the publicly available split in \cite{Song2015a,Wang_2016_CVPR},
the 19 most common categories are selected, consisting of 4,845 images
for training and 4,659 images for test.

\paragraph{Classifier and features}

Since we found that training linear SVM classifiers with the output
of the fully connected layer increases slightly performance, all the
following results use SVMs, if not specified.

\begin{table}
\caption{\label{tab:Comparisons-on-only}Comparison of depth on SUN RGB-D}
\begin{centering}
\begin{tabular}{>{\raggedright}p{0.16\columnwidth}cc}
\hline 
\multirow{1}{0.16\columnwidth}{} & \multirow{1}{*}{Method} & \multicolumn{1}{c}{Acc.(\%)}\tabularnewline
\hline 
\multirow{2}{0.16\columnwidth}{Proposed} & D-CNN  & 41.2\tabularnewline
 & D-CNN (wSVM)  & \textbf{42.4}\tabularnewline
\hline 
\multirow{4}{0.16\columnwidth}{State-of-the-art} & R-CNN+FV\cite{Wang_2016_CVPR} & 34.6\tabularnewline
 & FT-Places-CNN\cite{Wang_2016_CVPR} & 37.5\tabularnewline
 & FT-Places-CNN+SPP & 37.7\tabularnewline
 & FT-Places-CNN+SPP (wSVM) & 38.9\tabularnewline
\hline 
\end{tabular}
\par\end{centering}
\vspace{-0.5cm}
\end{table}
\begin{table*}
\caption{\label{tab:Comp-RGB-D}Comparisons of RGB-D data on SUN RGB-D}
\begin{centering}
\begin{tabular}{cc>{\centering}p{0.27\columnwidth}>{\centering}p{0.27\columnwidth}ccc}
\hline 
\multirow{2}{*}{} & \multirow{2}{*}{Method} & \multicolumn{2}{c}{CNN models} & \multicolumn{3}{c}{Accuracy (\%)}\tabularnewline
 &  & RGB & Depth & RGB & Depth & RGB-D\tabularnewline
\hline 
\multirow{3}{*}{Baseline} & Concate. & \multirow{1}{0.27\columnwidth}{Places-CNN} & Places-CNN & 35.4 & 30.9 & 39.1\tabularnewline
 & Concate. & \multirow{1}{0.27\columnwidth}{FT-Places-CNN} & FT-Places-CNN & 41.5 & 37.5 & 45.4\tabularnewline
 & Concate. (wSVM) & \multirow{1}{0.27\columnwidth}{FT-Places-CNN} & FT-Places-CNN & 42.7 & 38.7 & 46.9\tabularnewline
\hline 
\multirow{3}{*}{Proposed} & RGB-D-CNN & \multirow{1}{0.27\columnwidth}{FT-Places-CNN} & FT-Places-CNN & 41.5 & 37.5 & 48.5\tabularnewline
 & RGB-D-CNN & FT-Places-CNN & D-CNN & 41.5 & 41.2 & 50.9\tabularnewline
 & RGB-D-CNN (wSVM) & FT-Places-CNN & D-CNN & 42.7 & 42.4 & \textbf{52.4}\tabularnewline
\hline 
\multirow{2}{*}{State-of-the-art} & \cite{Zhu_2016_CVPR} & \multirow{1}{0.27\columnwidth}{FT-Places-CNN} & FT-Places-CNN & 40.4 & 36.5 & 41.5\tabularnewline
 & \cite{Wang_2016_CVPR} & \multirow{1}{0.27\columnwidth}{FT-Places-CNN + R-CNN} & FT-Places-CNN + R-CNN & 40.4 & 36.5 & 48.1\tabularnewline
\hline 
\end{tabular}
\par\end{centering}
\vspace{0cm}
\end{table*}
\begin{itemize}
\item (wSVM): this variant uses category-specific weights during SVM training
to compensate the imbalance in the training data. The weight $w=\{w_{1}...w_{K}\}$
of each category $k$ is computed as $w_{k}=\left(\frac{\min_{i\in K}N_{i}}{N_{k}}\right)^{p}$,
where $N_{k}$ is the number of training images of the $k_{th}$ category.
We select $p=2$ by cross validation. 
\end{itemize}

\paragraph{Evaluation metric}

Following \cite{Song2015a,Wang_2016_CVPR}, we report the average precision
over all scene classes for both datasets.

\subsection{Evaluations on SUN RGB-D}

\paragraph{Ablation study. }

We compare D-CNN and Alex-CNN on SUN RGB-D depth data. The outputs
of different layers are used as features to train the SVM classifiers.
We select 5 different models for comparison in Table~\ref{tab:Comparisons-of-different}.
We use the Alex-CNN architecture with only 3 bottom convolutional
layers for training from scratch. With Alex-CNN architecture, the
bottom layers (pool1 to conv3) trained from scratch perform better
than the transferred from Places-CNN, even though the top layers are
worse. Using weakly-supervised training on patches (WSP), the performance
is comparable to fine tuned Places-CNN for top layers and better for
bottom layers, with a smaller model and without relying on Places
data. D-CNN consistently achieves the best performance.

\paragraph{Comparisons with depth data}

We compare to related methods on depth recognition in Table~\ref{tab:Comparisons-on-only}.
For the fair comparison, we also implement SPP on Places-CNN for the
fine tuning. Our D-CNN outperforms FT-Places-CNN+SPP with 3.5\% in
accuracy. When both models using weighted SVM for training, our D-CNN
works even better.

\begin{table}
\setlength{\tabcolsep}{2pt} \renewcommand{\arraystretch}{1}\caption{\label{tab:Comparisons-on-NYUD2}Comparisons on NYUD2 in accuracy(\%)}
\begin{centering}
\begin{tabular}{c>{\centering}p{0.27\columnwidth}>{\centering}p{0.27\columnwidth}c}
\hline 
\multicolumn{1}{c}{} & \multicolumn{2}{c}{Features} & \multirow{2}{*}{Acc.}\tabularnewline
Method & RGB & Depth & \tabularnewline
\hline 
\multicolumn{4}{c}{Baseline methods}\tabularnewline
\hline 
RGB & \multirow{1}{0.27\columnwidth}{FT-Places-CNN} &  & 53.4\tabularnewline
Depth & \multirow{1}{0.27\columnwidth}{} & FT-Places-CNN & 51.8\tabularnewline
Concate. & \multirow{1}{0.27\columnwidth}{FT-Places-CNN} & FT-Places-CNN & 59.5\tabularnewline
\hline 
\multicolumn{4}{c}{Proposed methods}\tabularnewline
\hline 
Depth & \multirow{1}{0.27\columnwidth}{} & D-CNN & 56.4\tabularnewline
RGB-D-CNN & FT-Places-CNN & D-CNN & \textbf{65.8}\tabularnewline
\hline 
\multicolumn{4}{c}{State-of-the-art}\tabularnewline
\hline 
\cite{Gupta2014} & \multicolumn{2}{c}{Segmentation responses} & 45.4\tabularnewline
 \cite{Wang_2016_CVPR} & \multirow{1}{0.27\columnwidth}{FT-Places-CNN + R-CNN} & FT-Places-CNN + R-CNN & 63.9\tabularnewline
\hline 
\end{tabular}
\par\end{centering}
\vspace{0cm}
\end{table}

\paragraph{RGB-D fusion }

We compare to the state-of-the-art works \cite{Zhu_2016_CVPR,Wang_2016_CVPR}
of RGB-D indoor recognition in Table~\ref{tab:Comp-RGB-D}, where
a discriminative RGB-D fusion CNN is proposed in \cite{Zhu_2016_CVPR}
and a joint feature fusion of RGB-D, scene and objects is proposed
in \cite{Wang_2016_CVPR}. The proposed RGB-D-CNN outperforms the
RGB-D fusion method in \cite{Wang_2016_CVPR} with 2.7\% with linear
SVM, 4.3\% with weighted SVM, without including external training
of R-CNN \cite{GuptaECCV2014} as in that approach.

\subsection{Comparisons on NYUD2}

We compare our RGB-D-CNN to the state-of-the-art on NYUD2 in Table~\ref{tab:Comparisons-on-NYUD2}.
Gupta \textit{et al. \cite{Gupta2014} }propose to encode segmentation
responses as features for scene recognition. Our RGB-D-CNN largely
outperforms this work, where the segmentation is based on the hand-crafted
features. Comparing to RGB-D fusion in \cite{Wang_2016_CVPR}, we
achieves a gain of 1.9\% in accuracy, without including the R-CNN
models as in that work.

\section{Conclusion}

Transferring deep representations within the same modality (e.g. Places-CNN
fine tuned on SUN397) works well, since low-level patterns have similar
distributions, and bottom layers can be reused while adjusting the
more dataset-specific top layers. However, fine tuning is not that
effective in inter-modal transfer, such as Places-CNN to depth in
the HHA space, where low-level features require modality-specific
filters. In this paper, we focus on the bottom layers, because they
are more critical to represent depth data properly. By reducing the
number of parameters of the network, and using weakly supervised learning
over patches, the complexity of the model matches better the amount
of data available. This depth representation is not only more discriminative
than those fine tuned from Places-CNN but also when combined with
RGB features the gain is higher, showing that both are complementary.
Notice also, that we do not depend (for depth) on large datasets such
as Places.

\section*{Acknowledgment}

This work was supported in part by the National Natural Science Foundation
of China under 61322212, Grant 61532018, and Grant 61550110505, in
part by the National High Technology Research and Development 863
Program of China under Grant 2014AA015202, in part by the Beijing
Municipal Commission of Science and Technology under Grant D161100001816001,
in part by the Lenovo Outstanding Young Scientists Program, in part
by National Program for Special Support of Eminent Professionals and
National Program for Support of Top-notch Young Professionals. 

{\small{}\bibliographystyle{aaai}
\bibliography{AAAI17_v1}
}{\small \par}
\end{document}